%% file: main.tex
\documentclass[runningheads]{llncs}
\usepackage[T1]{fontenc}
\usepackage{wrapfig}
\usepackage{graphicx}
\usepackage{booktabs}
\usepackage{amsmath}
\usepackage{tabularx}
\usepackage{float}
\usepackage{multirow}
\usepackage{array}
\usepackage{spverbatim}
\usepackage{hyperref}
\usepackage{longtable}
\usepackage{xcolor}
\newcolumntype{S}{>{\raggedright\arraybackslash}p{0.23\textwidth}}

\newcommand{\bgcom}[1]{\textcolor[rgb]{0.87,0.28,0.00}{$<$\textbf{BG: }\textit{#1}$>$}}
\usepackage[utf8]{inputenc}

\begin{document}

\title{OntoGSN: An Ontology-Based Framework for Semantic Management and Extension of Assurance Cases}
\titlerunning{OntoGSN: Semantic Assurance Framework}

\author{Tomas Bueno Momcilovic\inst{1}\orcidID{0000-0003-4503-2244} 
\and Barbara Gallina \inst{2}\orcidID{0000-0002-6952-1053} 
\and Ingmar Kessler\inst{1}\orcidID{0000-0002-9945-7144}
\and Jule Hendricks\inst{1}\orcidID{0009-0004-2853-8408} 
\and Dian Balta\inst{1}\orcidID{0000-0001-8311-3227}}

\authorrunning{T. Bueno Momcilovic et al.}

\institute{
fortiss research institute, Munich, Germany \\
\email{(momcilovic,balta,ikessler)@fortiss.org}\\ \and
Mälardalen University, Västerås, Sweden \\
\email{barbara.gallina@mdu.se}
}

\maketitle

\begin{abstract} 
Assurance cases (ACs) are a common artifact for documenting and maintaining confidence in system properties such as safety or security. However, their underlying semantics are often implicit, tool-specific, and difficult to maintain as systems evolve. Existing tools largely provide siloed, document-centric graph editors, which deepens the gaps between AC, system assurance workflows, and the context-sensitive models of systems themselves. In this paper, we present OntoGSN, a three-layered, ontology-based framework for the semantic management and extension of ACs. OntoGSN aims to reduce semantic gaps across tools and domains by providing a unified representation and reusable components that support reasoning over both ACs and domain ontologies. At its core is a knowledge layer consisting of 1:1 OWL/SWRL formalization of Goal Structuring Notation (v3), with explicit provenance to each sentence of the normative text. The operations layer provides a library of SPARQL queries for context-sensitive CRUD operations, checks, rules and hooks over a client-side triple store. The interface layer is a prototypical web-based application that supports integration along three axes: ontologies; tool-specific data; and LLM-based pipelines. We evaluate OntoGSN using FAIR principles, the OOPS ontology pitfall scanner, and competency questions, and report early findings from case studies and engagements with practitioners.

\end{abstract}

\noindent \textbf{Resource type:} Ontology; Software
\newline \textbf{License:} CC-BY-4.0; MIT License
\newline \textbf{DOI}: \url{https://doi.org/10.5281/zenodo.17808331}
\newline \textbf{URL}: \url{https://w3id.org/OntoGSN/}

\input{contents/intro}

\input{contents/background}

\input{contents/relwork}

\input{contents/design}
\input{contents/artifact}

\input{contents/eval}

\input{contents/conclusion}

\begin{credits}
\subsubsection{\textit{Resource Availability Statement:}} Source code for the OntoGSN ontology, website, application and supporting tools, as well as documentation, are available on GitHub\footnote{\url{https://github.com/fortiss/OntoGSN}}. The ontology is published under the Creative Commons 4.0 CC-BY-SA license, the website is available on \url{https://w3id.org/OntoGSN}, and the ontology is hosted under \url{https://w3id.org/OntoGSN/ontology}.

\subsubsection{\textit{Use of AI tools:}} We have used ChatGPT models 4o, 5.0 and {5.1} ("instant" and "extended thinking" modes) for partial help in: clarifying our claims and improving how they are formulated; generating/formatting example data for our ontologies; formatting \texttt{BibTeX} citations; and generating JS/HTML code snippets for the workbench. No AI-generated text or code has been used without detailed reviews and testing from the authors.

\subsubsection{\ackname} We thank Yannick Landeck, Radouane Bouchekir, Tewodros Beyene, Laure Millet and Carmen Carlan for feedback. Additionally, we thank Will Franks from Adelard (ASCE), Laure Millet from Critical Systems Lab (Socrates) and Ewen Denney from NASA Ames Research Center (AdvoCATE) for providing access to their respective tools. This work was conducted using Protégé and was partially supported by financial and other means by the following research projects: DUCA (EU grant agreement 101086308), DiProLeA (German Federal Ministry of Education and Research, grant 02J19B120 ff), ROBIN at the Bavarian Research Institute for Digital Transformation, $\infty$ COMPASS (\#49 project at Software Center, Sweden), as well as our industrial partners in the FinComp project.
\bgcom{in case you need space, the first names of all these people could be just shortened by indicating only the first letter..}
\end{credits}

\bibliographystyle{splncs04}
\bibliography{contents/refs}

\input{contents/appendix}

\end{document}

%% file: contents/intro.tex
\section{Introduction}\label{sec:intro}

Assurance cases (ACs) are widely used artifacts for documenting, structuring, and maintaining confidence in properties of systems~\cite{acwg2021gsn}, such as the safety of rockets~\cite{whitman2020artemis} or the security of software applications~\cite{mohamad2021security}. They are typically represented as hierarchical argument graphs in a visual and textual interface between developers and auditors, to support their decisions about whether, and to what extent, such properties have been satisfied. Originally conceived as \emph{safety cases} for the certification of physical systems, they have since been applied more widely, including to more opaque targets such as the non-harmfulness of frontier large language models~\cite{goemans2024safety}.

However, to understand and evaluate each part of an AC, developers and auditors need to be deeply knowledgeable in its specific context. As the complexity of systems and properties increases, ACs can become very large and complex as well, making it difficult for any one person to build a mental model of the entire argument. To reduce the cognitive load, notations (e.g.~\cite{acwg2021gsn}) and guidelines (cf.~\cite{rushby2015understanding}) have emerged to provide the methodological support over the years, while individual communities developed patterns (cf.~\cite{ward2020pattern}) and tools (cf.~\cite{adelard2022asce}) for technical support. Yet, the ACs themselves and the linked artifacts are often not shared beyond individual organizations, leading each silo to develop its own interpretations of what constitutes \emph{sufficient} semantic context. In combination with increasingly dynamic and complex systems~\cite{asaadi2020dynamic}, these widening \textbf{semantic gaps} worsen the risks of misinterpretation and false confidence.

We present the OntoGSN framework as a means of bridging these semantic gaps between claims and the knowledge or context necessary to interpret them. Its primary goal is to facilitate the understanding and evaluation of ACs by providing the developers and auditors with mechanisms to manage and extend their semantics with domain ontologies, on top of a comprehensive formalization of the Goal Structuring Notation (GSN). OntoGSN consists of contributions of open artifacts across three layers:
\begin{enumerate}
    \item \textbf{Core}: A 1:1 representation of all normative sections of the GSN Community Standard v3~\cite{acwg2021gsn} in OWL, SWRL and SPARQL, with (to our knowledge) the most comprehensive provenance documentation, coverage of rules, and serializations adapted for common use cases.
    \item \textbf{Middleware}: A bundle of runnable and reusable JavaScript utilities and SPARQL queries, that form an explicit boundary between the ontological core and client applications, enabling standard CRUD\footnote{create, read, update and delete} and advanced operations over a client-side Oxigraph triplestore.
    \item \textbf{Interface}: A fully client-side, JavaScript-based interface with reusable components and ontology templates for displaying, extending and experimenting with coordinated views and editors over ACs and linked artifacts.
    \item \textbf{Open Artifacts}: A free and open-source release of all ontologies, queries and code, with detailed documentation, evaluated quality and persistent identifiers following the Semantic Web best practices. 
\end{enumerate}

The remainder of the paper is structured as follows: \nameref{sec:background} explains the need for semantic management and extension; \nameref{sec:relwork} contextualizes our contribution within the community; \nameref{sec:design} provides the methodological background; \nameref{sec:artifact} details the contributions; \nameref{sec:eval} expands with evaluation results and examples; and \nameref{sec:conclusion} summarizes the work with a future outlook.

\begin{wrapfigure}{r}{0.35\textwidth}
  \vspace{-20pt}
  \centering
  \includegraphics[width=0.4\textwidth]{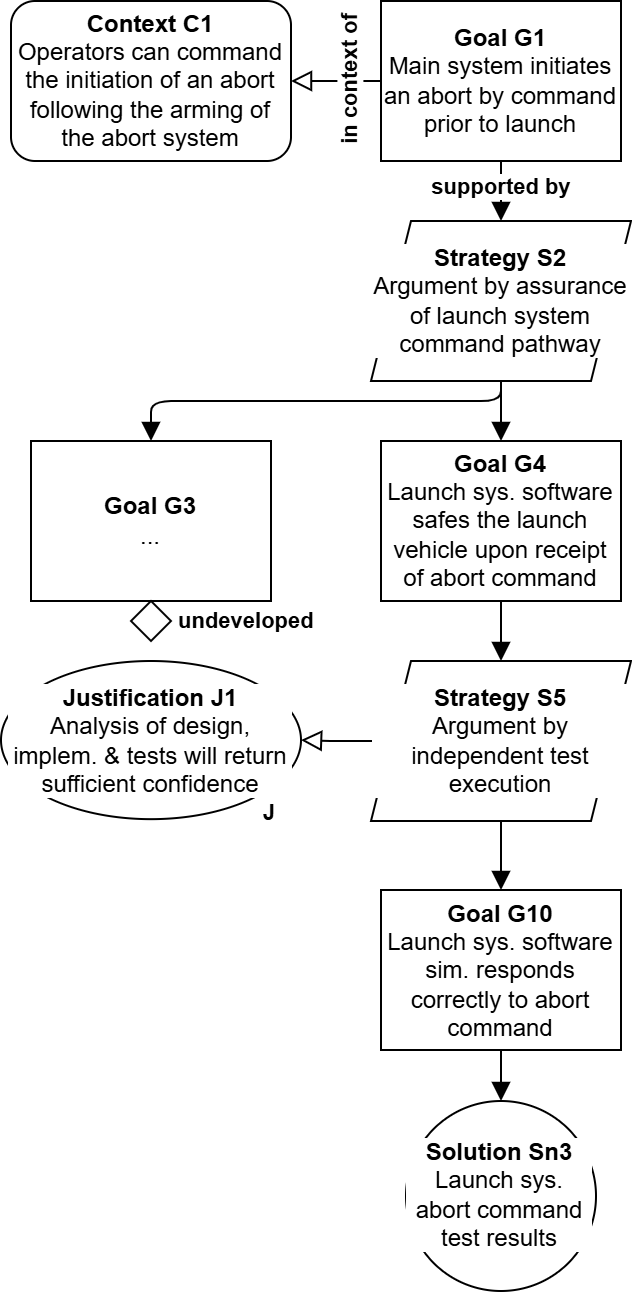}
  \caption{Example AC fragment for a rocket launch system, adapted from Whitman et al. (2020;~\cite{whitman2020artemis}) and paraphrased.}
  \label{fig:artemis_ac}
  \vspace{-15pt}
\end{wrapfigure}

%% file: contents/background.tex
\section{Background}\label{sec:background}

Assurance cases (ACs) are directed acyclic graphs that represent hierarchical arguments over claims and evidence. A typical assurance argument inverts the usual order of a natural-language argument: a root-level claim about some system property is stated first, and then decomposed into logical branches of supporting sub-claims, which in turn terminate in leaf nodes pointing to concrete evidence. A complete AC thus consists of both the argument structure and the associated evidence artifacts, which are supplied together to auditors.

ACs are most commonly authored and reviewed using graphical notations such as the Structured Assurance Case Metamodel~\cite{omg2023sacm}, or the Goal Structuring Notation~\cite{acwg2021gsn} (GSN). The metamodel of core GSN and its four extensions is recorded in the community standard, most recently version 3. At its core, claims are represented as \emph{goals} and evidence as \emph{solutions}, but four additional node types are introduced: \emph{contexts} for explanatory information; \emph{assumptions} for a priori expectations; \emph{justifications} for node rationale; and \emph{strategies} for the logic of goal decomposition.

A typical GSN argument (cf. Fig.~\ref{fig:artemis_ac}) includes a combination of node types along every branch. Each type is visually distinguishable by shape and identifier, and each node contains a single line of text. However, only goals and solutions are necessary components; if a simple branch can be \emph{sufficiently} understood alongside the supplied evidence, there is no imperative for more\footnote{For example, the normative part of the standard \cite{acwg2021gsn} states that an AC "will demonstrate that a given system is acceptably safe in a given context" (p. 10) and that "claims can only be asserted to be true in a specified context" (p. 22). However, it follows that "context elements \emph{can} be used in GSN to make this relationship clear" (p. 22; emphasis ours); under a stricter interpretation, this means that explicit context nodes are not necessary elements.}. Thus, the need for further semantics is determined by the developers and auditors.

One trend strongly indicates that there is a need for extended syntax and semantics. While Part 1 defines the formal, normative and minimal GSN, the assurance community is relying on additional non-normative conventions and seeking more advanced techniques for managing ACs. For example, notable general (cf., Part 2 of~\cite{acwg2021gsn};~\cite{acwg2021guidance}) and context-specific guidelines (cf.,~\cite{iaea2012safety},~\cite{ward2020pattern}) emphasize methods to deepen and improve the node contents\footnote{Arguably, the extensions of GSN are good examples of community needs or practices that have been recognized by the standard contributors and absorbed by the normative text.}. Popular tools (cf.~\nameref{sec:relwork}) that support GSN and other AC metamodels are increasingly adding features beyond visual or textual node editing, e.g., integration of data and automation ~\cite{asaadi2020dynamic}, formalization~\cite{bloomfield2022assurance20}, HTML documentation and attachments~\cite{adelard2022asce}, domain models~\cite{misra2024automated}, or integration of large language models~\cite{diemert2025llm}. Therefore, when such conventions or features become necessary or widely used, they function as the \emph{de facto} syntax and semantics, even if they are not formally part of the \emph{de jure} standard.

Nonetheless, implicit and non-normative practices cannot be easily added to the normative GSN formalization without consensus of its community and contributors. They are part of an ongoing discussion, whereby any new convention (cf., \cite{graydon2015formal}) or feature (cf., \cite{graydon2025llm,diemert2025llm}) may not be widely accepted or applicable, before it goes through a process of clarification, interpretation beyond the literal text, or voting. Therefore, they cannot be represented as a faithful and uncontroversial formalization of the standard, yet they are based on legitimate needs, and inextricably linked to GSN. We suggest that the benefits of the Semantic Web practices can help balance these contradictions.

%% file: contents/relwork.tex
\section{Related Work}\label{sec:relwork}

The use of ontologies for managing ACs is not a novel idea, but current implementations are limited and publicly unavailable. The Structured Assurance Case Metamodel v1.1~\cite{omg2015sacm}, for example, originally provided mechanisms to link external OWL domain ontologies. The ECSEL AMASS project~\cite{ruiz2016amass,vara2019amass,vara2020amass} was an open-source certification platform that supported ontologies based on \textit{Open Services for Lifecycle Collaboration}~\cite{gallina2016oslc}, which connected assurance cases with artifacts, pioneering the benefit of ontology-driven AC management (cf.~\cite{gallina2017oslc}). Denney (2020)~\cite{denney2020ontology} explored the benefits of domain-specific ontologies in their AdvoCATE tool~\cite{denney2018advocate}, which supports GSN cases. More recently, MISRA (2024)~\cite{misra2024automated} defined custom domain ontologies within SysML to shape assurance cases for automated driving. Unfortunately, the results of these related initiatives have either been deprecated or are publicly unavailable (and therefore, difficult to work with or extend).

Software such as ASCE~\cite{adelard2022asce} and AdvoCATE~\cite{denney2020ontology} are widely used by assurance experts, and provide useful graphical user interfaces. However, similar issues emerge. These tools are closed-source and often proprietary; being specifically geared towards ACs, their adoption by a wider audience and thereby extensibility may be limited. Additionally, the logic and coverage of performed evaluations (i.e., tests or rules) in some of these tools is unclear, minimal and/or different (e.g., extended GSN of \cite{denney2018advocate}).

Related academic resources show promising new features that motivate extension. Castillo-Effen et al.~\cite{castillo2024digital}, for example, introduce an algorithmic framework and domain-specific language for managing ACs in the aerospace domain with data; Beyene \& Carlan \cite{beyene2021cybergsn} similarly introduce a the CyberGSN language for enabling features such as digital signatures for cybersecurity ACs. Saswata et al. \cite{paul2024triplestore} provide a triplestore-based framework (i.e., graph database) with patterns that can be instantiated into ACs using domain ontologies. Others, such as Asaadi et al. \cite{asaadi2020dynamic}, give architectural and development lifecycle recommendations for data-driven applications that are tool-agnostic.

We see the positioning of OntoGSN as highly complementary in following ways. First, the related work contributes with focused representations of GSN that fulfill a particular purpose. We instead provide a complete, comprehensive, and source-driven formalization of GSN that can be integrated into any framework, distilled into smaller models for a particular purpose, and aligned with other AC metamodels. Second, the OntoGSN ontology and workbench are public, free and open-source, thereby allowing easy extension of facts and rules from the domain, but also testing of ACs before deciding for proprietary or permissioned tools. Third, this contribution does not aim to substitute state-of-the-art graphical user interfaces, frameworks and pattern libraries; it is designed as a complementary middleware that fills the gap between multiple frameworks. Lastly, the aim of OntoGSN is to align GSN and non-GSN metamodels, extensions and less-represented properties in need of assurance (e.g., adversarial robustness) with normative and community-based practices.

%% file: contents/design.tex
\section{Design}\label{sec:design}

\subsection{Method}
The ontology was designed following the guidelines of the NeOn methodology~\cite{suarez2012neon} for the (re)use of non-ontological resources and existing ontologies. The GSN specification, as crystallized in the GSN Community Standard v3 \cite{acwg2021gsn}, represents one such non-ontological resource. The standard is the result of many years of iterative community efforts to standardize the documentation of arguments in a structured way. Therefore, early on, it was decided that the normative parts (i.e., Part 1, alongside Part 0\footnote{Although it is classified as "informative" in the Introduction (page 9, \cite{acwg2021gsn}), Part 0 contains definitions and descriptions of rules with normative qualifiers ("must")} and the Glossary) of the standard represent the ground truth, to which any interpretations, guidelines (e.g., Part 2), and supplementary specifications can be added in the future. 

Each sentence of the standard was broken down into elements and analyzed with respect to its contribution to the metamodel of GSN. There were two main activities: creating the taxonomy of classes and properties (i.e., the TBox), and creating the rules governing the properties between individuals of classes. Regarding the taxonomy, each sentence of the standard is parsed with the goal of translating the concepts and their relations into semantic triples (i.e., subject-predicate-object statements). This also includes possibilities denoted by qualifiers such as "can". Regarding the rules, the sentences that place conditions or restrictions on the identified elements of triples are translated into logical statements. These sentences were identified based on qualifiers such as "should" or "must", as opposed to qualifiers such as "may".

Given that concepts can be represented in multiple ways, the process of translating text to the ontology involves some degree of interpretation. To mitigate bias, the entire document was parsed three times, each \textit{text-to-ontology} mapping was traced, and the mappings were reviewed by all co-authors and additional early reviewers. Following the "design rationale" paradigm~\cite{potts1988rationale}, all text snippets were included side-by-side with the corresponding ontology statements and rationales in a tabular design document. The document thus served two purposes: 1) provide conceptual version control in a format that allows reviewers to easily (in)validate the mappings; and 2) ensure the ontology designers and reviewers have comprehensively covered the standard.

\subsection{Technical Details}
The ontology was created using the Protégé v5.6.3 ontology editor~\cite{musen2015protege}, following the Web Ontology Language (OWL 2) standard~\cite{w3c2012owl2}. Annotation and data properties were imported from the following foundational ontologies and vocabularies: Resource Description Framework Schema (RDFS)~\cite{w3c2014rdf}, XML Schema Definition Language (XSD)~\cite{w3c2012xml}, Dublin Core (DC)~\cite{dublin2020dcmi}, Schema.org~\cite{schema2025schema}, and Simple Knowledge Organization System (SKOS)~\cite{miles2009skos}. Reasoning is based on Semantic Web Rule Language (SWRL)~\cite{w3c2004swrl} rules and OWL axioms, which can be executed with supported rule engines (e.g., Pellet~\cite{sirin2007pellet} or Drools~\cite{jboss2025drools}). In this first version of OntoGSN, SWRL provides the simplicity, decidability, and reasoning support that covers most GSN rules, but SPARQL substitutes are provided.

\subsection{Competency Questions}

Following the example of~\cite{berardinis2023polifonia}, we drafted exemplary personas based on the stakeholders, which we perceive to be the intended beneficiaries of our framework: assurance engineers (i.e., internal auditors), domain experts (i.e., developers), and (external) auditors. Each persona is associated with five competency questions from an ongoing case study on assurance of adversarial robustness of commercial LLMs (cf.~\cite{bueno2024assurance}). For more information, consult \nameref{appendix:cq}.

We imagine the personas having particular expertise and interests. Assurance engineers want to build a strong GSN assurance case and make sure that the overall case is valid in the face of changes. Domain experts want to ensure that their assurance-relevant controls in applications (e.g., AI systems) are still valid (e.g., adversarially robust). Finally, auditors want to verify and validate the controls effectively, and provide targeted evaluations.

%% file: contents/artifact.tex
\section{The OntoGSN Framework}\label{sec:artifact}

\subsection{Ontological Core}

The OntoGSN core comprises 227 RDF/OWL axioms and 51 SWRL rules (cf. Table \ref{tab:gsn-assertions}), each annotated with \texttt{coreOrExtension} to indicate its source. The ontology is published under the W3ID~\cite{w3c2013w3id} namespace \url{https://w3id.org/OntoGSN/ontology} and the prefix \textbf{gsn} is registered with prefix.cc~\cite{cyganiak2010prefixcc}. Files and documentation are publicly available on GitHub (\url{github.com/fortiss/OntoGSN/}) and on via the dedicated website (\url{w3id.org/OntoGSN/}).

\begin{table}[!htbp]
  \vspace{-10pt}
  \centering
  \begin{tabular}{lccccc}
    & & \multicolumn{3}{c}{Number of Axioms (Protégé)} & \\
    \cmidrule(lr){3-5}
    Source & Pages | & Declaration | & Logical: OWL | & Logical: SWRL | & Total \\
    \midrule
    Ontology Metadata & {-} & 33 & {-} & {-} & 33 \\
    Core GSN$^{*}$ & 16{-}25 & 24 & 57 & 18 & 99 \\
    Argument Pattern Ext. & 26{-}31 & 26 & 25 & 3 & 54 \\
    Modular Ext. & 32{-}45 & 10 & 24 & 17 & 51 \\
    Confidence Argument Ext. & 46{-}49 & 3 & 12 & 6 & 21 \\
    Dialectic Ext. & 50{-}53 & 4 & 13 & 7 & 20 \\
    \midrule
    Total & & 100 & 127 & 51 & 278 \\
    \addlinespace[0.3em]
    \multicolumn{6}{l}{\footnotesize $^{*}$Includes Part 0, Glossary, and Appendices.} \\
  \end{tabular}
  \caption{Axiom counts in the ontology, excluding annotations.}
  \label{tab:gsn-assertions}
\end{table}

\begin{figure}[!htbp]
  \vspace{-20pt}
  \centering
  \includegraphics[width=\linewidth]{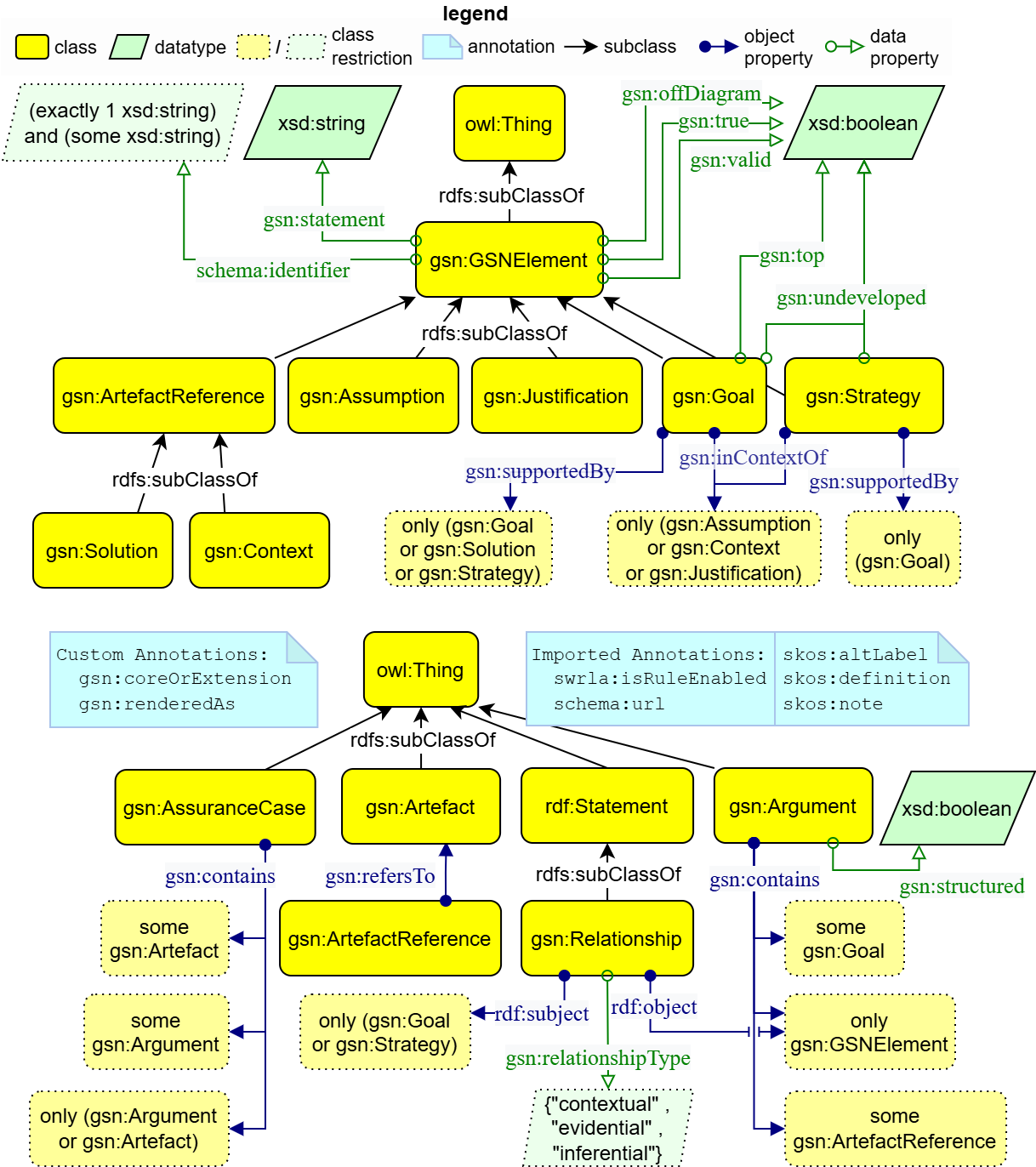}
  \caption{Graffoo~\cite{falco2014graffoo} diagram of Core GSN.}
  \label{fig:core1}
\end{figure}

\begin{sloppypar}
Fig.~\ref{fig:core1} shows the backbone of OntoGSN. The class \texttt{GSNElement} covers main node types: \texttt{Goal}, \texttt{Strategy}, \texttt{Assumption}, \texttt{Justification}, and \texttt{ArtefactReference} (covering \texttt{Context} and \texttt{Solution}). It collects common logical axioms, such as mandatory \texttt{schema:identifier} and \texttt{statement}, and serves as a domain for normative and implicit Boolean data properties (e.g., \texttt{valid}, \texttt{true}). \texttt{GSNElement}s primarily interface with the central node type \texttt{Goal} via the \texttt{supportedBy} and  \texttt{inContextOf} object properties. 

To structure complete ACs, four container classes are introduced: \texttt{Argument} containing at least one \texttt{Goal} and \texttt{ArtefactReference}; \texttt{Artefact} denoting the actual evidence file(s); \texttt{AssuranceCase} containing at least one \texttt{Argument} and \texttt{Artefact}; and \texttt{Relationship} (a subclass of \texttt{rdf:Statement}) reifying semantic triples to enable reasoning over individual relations.
\end{sloppypar}

\begin{figure}[!htbp]
  \vspace{-10pt}
  \centering
  \includegraphics[width=\linewidth]{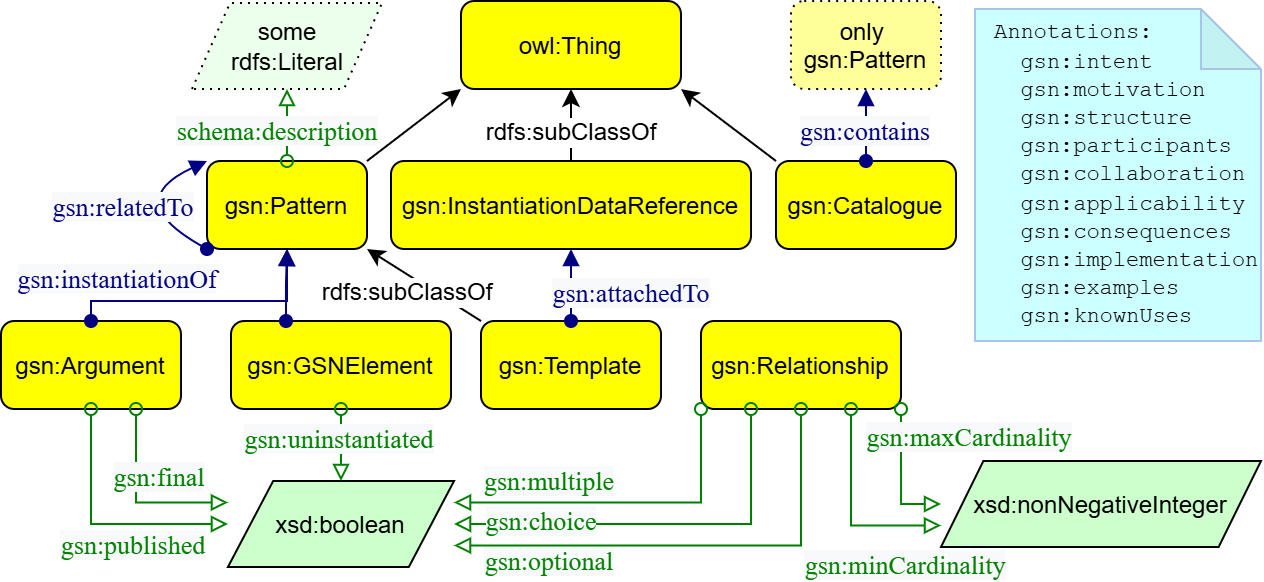}
  \caption{Argument Pattern Extension}
  \label{fig:patterns}
  \vspace{5pt}
  \includegraphics[width=0.95\linewidth]{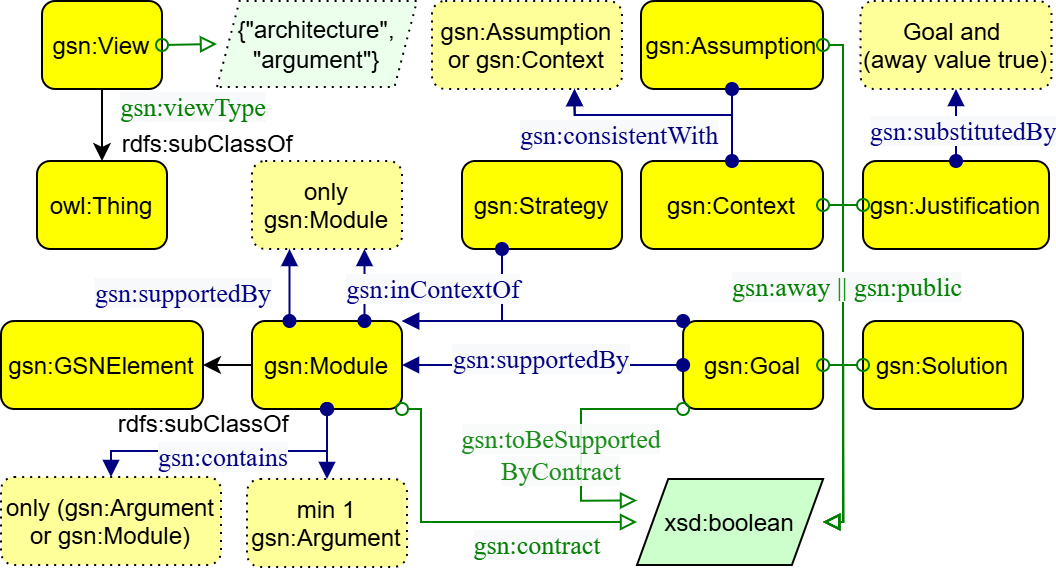}
  \caption{Modular Extension}
  \label{fig:modular}
\end{figure}

\begin{sloppypar}
Fig.~\ref{fig:patterns} introduces reusable \texttt{Pattern}s of \texttt{Argument}s and \texttt{GSNElement}s. Each pattern is documented via a small annotation schema and may be \texttt{relatedTo} other patterns. \texttt{Catalogue}s collect patterns, while \texttt{Template}s are special pattern instances \texttt{attachedTo} instantiation data (\texttt{InstantiationDataReference}). \texttt{Relationship}s with patterns as subjects encode cardinality constraints and instantiation indicators (\texttt{choice}, \texttt{multiple}, \texttt{optional}), while status flags (\texttt{published}, \texttt{final}, \texttt{uninstantiated}) describe the state of pattern instances.

Fig.~\ref{fig:modular} captures modularization of large ACs in logical partitions (i.e., \texttt{Module}s) and visual projections (i.e., \texttt{View}s). Both can \texttt{contain} one or more \texttt{Module}s or \texttt{Argument}s. The \texttt{viewType} \textit{argument} shows individual argument elements, whereas \texttt{architecture} shows aggregated relations between elements as relations between \texttt{Module}s. Additional flags (e.g., \texttt{gsn:public} for confidentiality) and rules (e.g., \texttt{consistentWith} checks between contexts) support information management.
\end{sloppypar}

\begin{figure}[!htbp]
  \vspace{-10pt}
  \centering
  \includegraphics[width=0.95\linewidth]{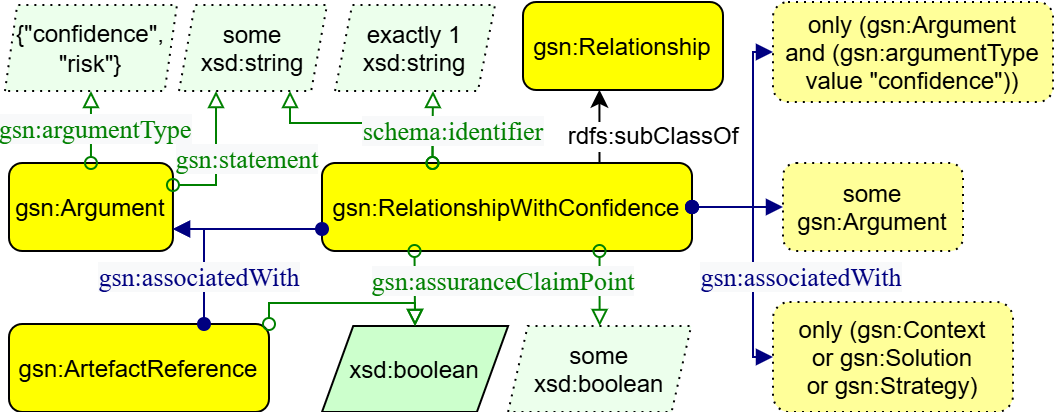}
  \caption{Confidence Argument Extension}
  \label{fig:confidence}
  \vspace{5pt}
  \includegraphics[width=0.9\linewidth]{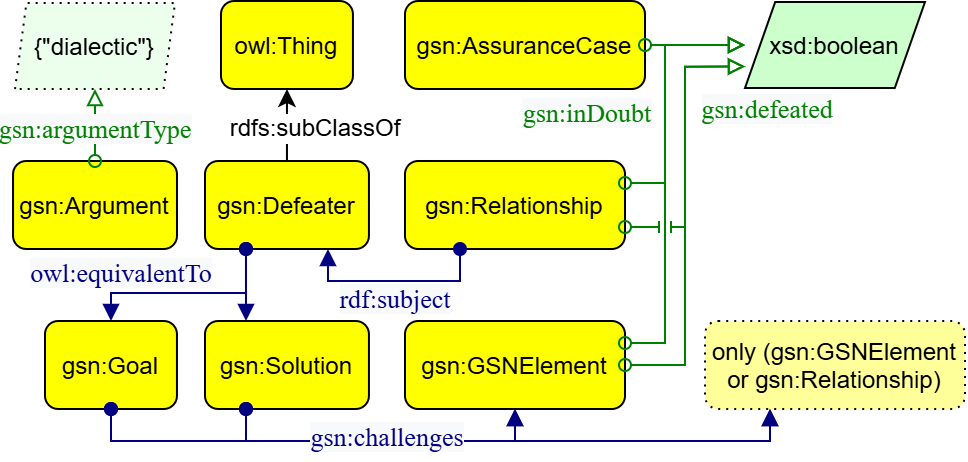}
  \caption{Dialectic extension}
  \label{fig:dialectic}
\end{figure}

\begin{sloppypar}
Fig.~\ref{fig:confidence} formalizes the expression of \textit{confidence arguments} about elements of \textit{risk arguments} (i.e., regular assurance claims) via \texttt{assuranceClaimPoint}s. To support reasoning, the \texttt{associatedWith} relation is reified as \texttt{RelationshipWithConfidence}, which inherits the axioms of \texttt{Relationship} while enabling.

Finally, Fig.~\ref{fig:dialectic} defines \texttt{Defeater}s, i.e., any \texttt{Goal} or \texttt{Solution} that \texttt{challenges} another element or a relation. When challenged, elements are either flagged as \texttt{inDoubt} (requiring review) or as \texttt{defeated} (requiring revision), depending on the strength of the challenge.
\end{sloppypar}

Each section of GSN provides implicit and explicit logic for constraints, propagation, and flagging; we encode these as SWRL rules and provide SPARQL equivalents. The first group of rules classify and reify relationships between GSN elements, deriving explicit Relationship individuals, marking top-level goals, and flagging conflicting triples as invalid. They also propagate true and valid flags along the graph, reflecting the impact of invalid assumptions, conflicting evidence, or overlapping contexts. The second group lifts these semantics to modules and templates (i.e., for improved information management): undeveloped or finalized instantiations, away nodes, cross-module links, identifier uniqueness, etc. The third group maintain relations relevant to confidence arguments, and introduces propagation through dialectic flags (“in doubt” or “defeated”). For more information, consult \nameref{appendix:swrl}.

\subsection{Middleware}

The middleware is a browser-based HTML/JS orchestration layer that sits between the OntoGSN knowledge base (GSN ontology with additional domain and AC ontologies populated with examples) and all visual components of the workbench. It initialises an Oxigraph~\cite{pellissier2023oxigraph} triplestore in the browser, loads the bundled Turtle ontologies, and exposes a single query API; the interface views can then call template or executable SPARQL queries with paths to dedicated files. The middleware decides whether the query is a SELECT or UPDATE, executes it against the store, and hands the results back in a uniform row format. On top of simple execution, it also adds utilities for the workbench interfaces, such as managing “overlays” for the graph (e.g., rule-triggered nodes) and artefacts, so that different views can highlight or group the same resources without duplicating logic.

From a developer’s perspective, this middleware is the backend of the small OntoGSN workbench: adding new behaviour means adding a new \texttt{.sparql} file and, optionally, a small amount of wiring to a button or rule checkbox. The JS code itself remains largely generic, and does not hard-code any domain- or case-specific assumptions. This separation makes it easy to swap in different assurance cases, domain models, or rulesets.

\subsection{Interface}

The interface presents the workbench frontend machinery for an interactive OntoGSN playground, aimed at users who want to explore ACs or features rather than develop tooling. It is organised as a two-pane layout: the left pane offers textual and tabular access to the case (a SPARQL-backed table view, a simple SPARQL editor for templated CRUD operations, and a document view that renders Markdown assurance narratives), while the right pane provides multiple coordinated visualisations. Users can switch between the views with buttons: “Tree View” of the argument graph; a “Layer View” that flattens goals as layered controls; or a “Model View” that renders an ontology-based 3D car model whose parts are linked to the underlying AC. For visual examples, see \nameref{label} and visit the interactive playground (\url{https://fortiss.github.io/OntoGSN/playground/}).

%% file: contents/eval.tex
\section{Assessment and Discussion}\label{sec:eval}

OntoGSN was checked at schema- and instance-levels. Regarding schema, using the OOPS!~\cite{poveda2014oops} service, which flagged: missing annotations, domain or range declarations, and inverse relationships, recursive definitions; and untyped classes. Missing assertions and recursion have been addressed, but the issue of untyped classes remains due to how SWRL is implemented in Protégé. Regarding instances, we verified consistency by loading a representative ABox into the Pellet OWL-DL reasoner, confirming that classes remained satisfiable and no logical contradictions arose in inferred class memberships or property assertions. Further, SWRL rules were additionally tested with the Drools reasoner.

\subsection{Examples}

The first Playground example models an assurance case for adversarial robustness of a large language model against jailbreaks, prompt-injection and data-exfiltration attacks. OntoGSN is used to encode the top-level robustness claims, their decomposition into defence-in-depth strategies, and the supporting evidence modules (e.g., benchmarks, red-team tests, filtering and monitoring components). A separate security ontology captures attack types, capabilities and defence mechanisms, which are linked to GSN goals and solutions so that properties such as coverage, overlap and gaps can be queried. Additional OntoGSN rules propagate successful attacks and misconfigurations through the argument graph as defeaters, highlight invalid assumptions, and track how challenges in one module undermine confidence in others, thereby illustrating modular, dialectic reasoning over a security-focused assurance case.

The second example focuses on static structural load safety for a small demonstration vehicle with a roof rack. A Schema.org-inspired car ontology specifies parts, dimensions and weight limits for the vehicle and its roof-mounted luggage, while an OntoGSN argument links these domain facts to claims about allowable payload and roof load in different driving scenarios. OntoGSN metadata ties individual goals and contexts to specific car components, and rules check consistency between stated limits and current loading configurations; when a counterfactual overloaded scenario is activated, violations are reified as defeaters and propagated through the assurance case. This example therefore demonstrates how OntoGSN can connect a relatively simple physical domain model with an assurance argument, and how rule-based propagation makes inconsistencies and unsafe configurations immediately visible in both the graph and the accompanying narrative/model views.

\subsection{Evaluating on Competency Questions}


The competency questions in Appendix \ref{appendix:cq} were formalized as SPARQL queries; all are available in the GitHub repository. For instance, the question
\begin{quote}
  \emph{Which Goals or Strategies are affected by the Defeater “Jailbreak”?}
\end{quote}
is realized by selecting all \texttt{gsn:Defeater} instances whose \texttt{gsn:statement} contains the literal “jailbreak” and then following the \texttt{gsn:challenges} property to retrieve the related \texttt{gsn:Goal} or \texttt{gsn:Strategy} resources. Likewise, to answer
\begin{quote}
  \emph{Which bottom-level Goals in the “Robustness” argument lack any supporting Solution?}
\end{quote}
one locates the \texttt{gsn:Argument} whose \texttt{gsn:statement} mentions “robustness”, traverses its \texttt{gsn:contains+} hierarchy to leaf \texttt{gsn:Goal} instances (i.e.\ those without further \texttt{gsn:contains} children), and filters out any goals that appear in a \texttt{gsn:supportedBy} relationship. A third example,
\begin{quote}
  \emph{Which Solution instances were added more than 180 days ago?}
\end{quote}
simply filters all \texttt{gsn:Solution} resources by comparing their \texttt{gsn:published} date‐time against a computed cutoff (e.g.\ “2024-11-14T00:00:00Z”). In each case, the OWL ontology’s classes, properties, and transitive paths directly inform the SPARQL patterns, guaranteeing that queries faithfully satisfy the original competency questions.

\subsection{Contribution}

OntoGSN provides a formal, extensible, interoperable core ontology and workbench for GSN. We argue that it represents a valuable contribution to the assurance community, that complements existing tools and advances the state of practice for managing and extending the semantics of ACs. Similarly, it introduces a critical but historically non-semantic domain to the Semantic Web community, demonstrating that OWL, SWRL , SPARQL, and graph-based reasoning are exceptionally well-suited for representing, analyzing, and automating structured arguments. This expands the Semantic Web’s footprint into safety- and security-critical engineering, where high-quality, explainable reasoning is urgently needed. We also believe this creates a clear path for members interested in both communities, while satisfying the needs of primary non-expert beneficiaries of ACs and Semantic Web technologies (i.e., developers and auditors).

\subsection{Further Evaluation, Maintenance and Adoption Strategy}

As a novel contribution, OntoGSN is being evaluated in ongoing review cycles across different communities. First, we are actively extending and evaluating OntoGSN in two case studies: assurance of IT security and compliance in the financial domain; and assurance of adversarial robustness of commercial LLMs. Second, we have disseminated OntoGSN with the members and leaders of the SCSC Assurance Case Working Group (i.e., the creators, contributors and maintainers of the GSN standard) and the Ontology Working Group, who directly expressed interest. Third, we are discussing integration of the OntoGSN ontology and/or workbench with existing tools (ASCE~\cite{adelard2022asce} and Socrates~\cite{csl2025socrates}), and use cases of interested contributors. Fourth, we have directly engaged ontology engineers and domain experts, who provided their comments with respect to ontology quality and usability. 

Our maintenance strategy is based on best practices of version control and open-source projects on Github, following the practices of authors of similar ontologies and frameworks (cf.~\cite{berardinis2023polifonia}). To promote contributions, we invite and engage practitioners to add issues to our repository's issue tracker. For wider adoption, we provide and continue to expand tutorials and user support.

\subsection{Limitations}
Several limitations remain in the current version of OntoGSN. First, we do not verify the syntactic correctness or semantic validity of \texttt{gsn:statement}s. Second, punning of object properties as classes to fit them under \texttt{rdf:predicate} in Relationships is not supported due to SWRL conflicts. Third, rules and axioms are not annotated according to their source (\texttt{gsn:coreOrExtension}) due to a known Protégé error, and their filtering by source is therefore currently disabled. Finally, while many SWRL rules can (and perhaps, should) be expressed as SPARQL queries~\cite{prudhommeaux2013sparql} or SHACL constraints~\cite{w3c2017shacl}, this translation effort has not been pursued in this version.

%% file: contents/conclusion.tex
\section{Conclusion and Future Work}\label{sec:conclusion}

OntoGSN addresses the persistent knowledge‐management bottleneck in AC development by providing a machine‐readable, standards-compliant, and publicly available formalization of GSN, accompanied by middleware for automated population, querying, and consistency checking. Through an ontology, extensive documentation, and supporting tools, OntoGSN enables engineers to construct, evolve, and evaluate cases with reduced manual effort. Our evaluation - encompassing FAIR principles, OOPS! pitfall analysis, competency‐question validation, and community feedback - and the workbench exemplify how semantic ACs can be realized.

Future work will include three directions: extending tool integrations, broadening framework compatibility, and supporting domain-specific use cases. On the tooling side, one aim will be to integrate OntoGSN with additional AC tools (e.g., AdvoCATE~\cite{denney2018advocate} or CertWare~\cite{barry2011certware}) and provide support for usage with Python. On the framework side, we plan to formalize and align our approach with existing ontologies and methodologies for dynamic AC management, with the aim to align with the Argument Model Ontology~\cite{vitali2011amo}, Assurance 2.0~\cite{bloomfield2022assurance20} and other notations (e.g., SACM \cite{omg2023sacm}). On the domain side, we aim to improve the interoperability and applicability of OntoGSN for diverse communities, environments and roles.


%% file: contents/appendix.tex
\newpage
\section{Appendix}

\subsection{Appendix 1: Human-Readable SWRL Rules}
\label{appendix:swrl}
\input{tables/swrl_rules}

\newpage

\subsection{Appendix 2: Interface Panels}

\begin{figure}[!htbp]
  \centering
  \includegraphics[width=\linewidth]{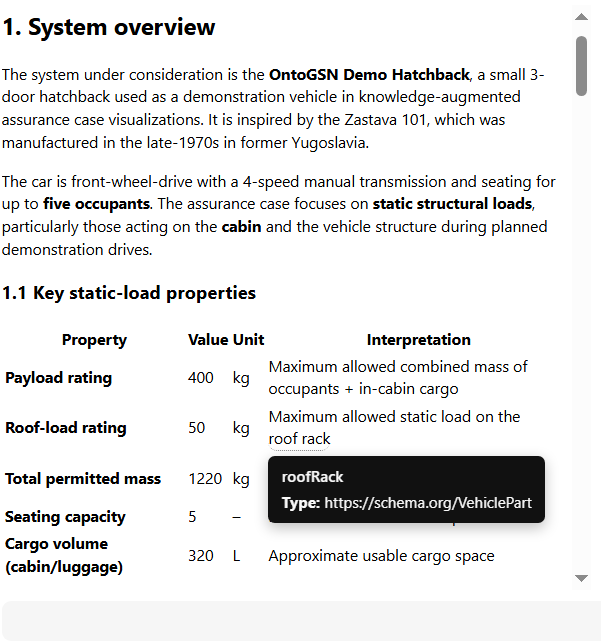}
  \caption{Document panel}
  \label{fig:interface_document}
\end{figure}

\begin{figure}[!htbp]
  \centering
  \includegraphics[width=\linewidth]{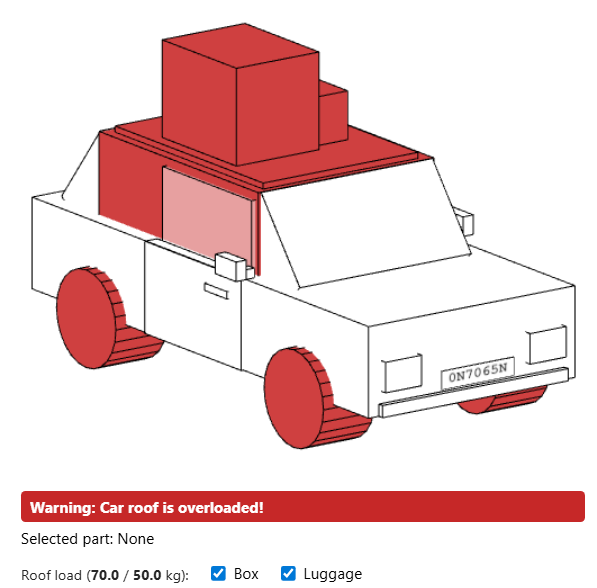}
  \caption{Model panel}
  \label{fig:interface_model}
\end{figure}

\begin{figure}[!htbp]
  \centering
  \includegraphics[width=\linewidth]{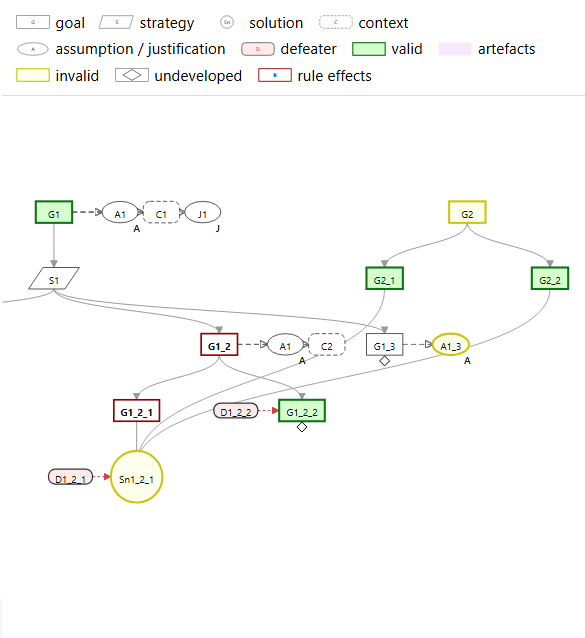}
  \caption{Graph panel}
  \label{fig:interface_graph}
\end{figure}

\newpage 
\subsection{Appendix 3: Competency Questions}
\label{appendix:cq}

\input{tables/competency_questions.tex}

%% file: tables/swrl_rules.tex
Choice of variables (e.g., \textbf{A}, \textbf{M}, \textbf{R1}, ...)  is arbitrary.

\small
\begin{longtable}{@{}%
  >{\raggedright\arraybackslash}p{7.3cm}%
  >{\raggedright\arraybackslash}p{6.7cm}%
@{}}

\toprule
\textbf{Antecedent (IF)} & \textbf{Consequent (THEN)} \\
\midrule

\textbf{A} is an \texttt{Assumption}
\newline \textbf{A} is \underline{not} \textit{valid} 
\newline \textbf{B} is in context of \textbf{A}
  & \textbf{B} is \underline{not} \textit{valid} 
\\ \\

\textbf{R} has subject \textbf{A}
\newline \textbf{A} is in context of \textbf{B}
\newline \textbf{R} has object \textbf{B}
  & \textbf{R} is a \textit{contextual} relationship 
\\ \\

\textbf{B} is a \texttt{Solution}
\newline \textbf{A} is supported by \textbf{B}
\newline \textbf{R} has subject \textbf{A}
\newline \textbf{R} has object \textbf{B}
  & \textbf{R} is an \textit{evidential} relationship
\\ \\

\textbf{B} is a \texttt{Strategy}
\newline \textbf{A} is supported by \textbf{B}
\newline \textbf{R} has subject \textbf{A}
\newline \textbf{R} has object \textbf{B}
  & \textbf{R} is an \textit{inferential} relationship
\\ \\

\textbf{B} is a \texttt{Goal}
\newline \textbf{A} is supported by \textbf{B}
\newline \textbf{R} has subject \textbf{A}
\newline \textbf{R} has object \textbf{B}
  & \textbf{R} is an \textit{inferential} relationship
\\ \\

\textbf{X} is supported by \textbf{A}
\newline \textbf{A} is supported by \textbf{B}
\newline \textbf{A} is a \texttt{Goal}
  & \textbf{A} is \underline{not} a \textit{top-level goal}
\\ \\

\textbf{R1} has subject \textbf{A}, predicate \textbf{P}, and object \textbf{B}
\newline \textbf{R2} has subject \textbf{B}, predicate \textbf{P}, and object \textbf{C}
\newline \textbf{R3} has subject \textbf{C}, predicate \textbf{P}, and object \textbf{A}
  & \textbf{R1} is \underline{not} \textit{valid}
\newline \textbf{R2} is \underline{not} \textit{valid}
\newline \textbf{R3} is \underline{not} \textit{valid}
\\ \\

\textbf{A} is \underline{not} \textit{true}
\newline \textbf{B} is \textit{true}
\newline \textbf{B} is supported by \textbf{A}
\newline \textbf{A} is a \texttt{Goal}
  & \textbf{B} is \underline{not} \textit{valid}
\\ \\

\textbf{C} is \textit{true}
\newline \textbf{B} is \textit{true}
\newline \textbf{C} is supported by \textbf{B} and \textbf{A}
\newline \textbf{A} and \textbf{B} are \texttt{Goal}s
  & \textbf{C} is \underline{not} \textit{valid}
\\ \\

\textbf{A} is a \texttt{Solution}
\newline \textbf{A} is supported by \textbf{B}
\newline \textbf{A} is \textit{true}
  & \textbf{B} becomes \textit{true}
\\ \\

\textbf{A} is a \texttt{Solution}
\newline \textbf{A} is \underline{not} \textit{true}
\newline \textbf{A} is supported by \textbf{B}
  & \textbf{B} becomes \underline{not} \textit{true}
\\ \\

\textbf{B} is \underline{not} \textit{valid}
\newline \textbf{A} is \textit{valid}
\newline \textbf{C} is supported by \textbf{B} and \textbf{A}
  & \textbf{C} is \underline{not} \textit{valid}
\\ \\

\textbf{A} and \textbf{B} are \texttt{Solution}s
\newline \textbf{B} is \underline{not} \textit{true}
\newline \textbf{B} supports \textbf{C}
\newline \textbf{A} is \textit{true}
\newline \textbf{A} supports \textbf{C}
  & \textbf{C} is \underline{not} \textit{true}
\\ \\

\textbf{R} is \underline{not} \textit{true}
\newline \textbf{R} has subject \textbf{A} and object \textbf{B}
\newline \textbf{A} is in context of \textbf{B}
\newline \textbf{R} is a \texttt{Relationship}
  & \textbf{A} is \underline{not} \textit{true}
\\ \\

\textbf{A} is an \texttt{Assumption}
\newline \textbf{A} is \underline{not} \textit{true}
\newline \textbf{B} is in context of \textbf{A}
  & \textbf{B} is \underline{not} \textit{true}
\\ \\

\textbf{B} is supported by \textbf{C}
\newline \textbf{B} is in context of \textbf{A}
\newline \textbf{A} is an \texttt{Assumption}
\newline \textbf{C} is a \texttt{Goal}
  & \textbf{C} is in context of \textbf{A}
\\ \\

\textbf{A} is supported by \textbf{B}
  & Create \textbf{R}
\newline \textbf{R} has subject \textbf{A} and object \textbf{B}
\\ \\

\textbf{A} is in context of \textbf{B}
  & Create \textbf{R}
\newline \textbf{R} has subject \textbf{A} and object \textbf{B}
\\ \\

\textbf{B} is an instantiation of \texttt{Pattern} \textbf{C} inside container \textbf{A}
\newline \textbf{A} is marked final
\newline \textbf{A} contains both \textbf{B} and \textbf{C}
\newline \textbf{C} is a \texttt{Pattern}
  & \textbf{A} is \underline{not} \textit{valid}
\\ \\

\textbf{A} is a \texttt{Template}
\newline there is an \textit{undeveloped} Relation \textbf{R} whose subject is \textbf{A}
\newline \textbf{A} is published
  & \textbf{A} is \underline{not} \textit{valid}
\newline \textbf{R} is \underline{not} \textit{valid}
\\ \\

\textbf{A} is a \texttt{Template}
\newline there is an \textit{undeveloped} Relation \textbf{R} whose subject is \textbf{B}
\newline \textbf{A} is published and contains \textbf{B}
  & \textbf{A} is \underline{not} \textit{valid}
\newline \textbf{R} is \underline{not} \textit{valid}
\\ \\

\textbf{M} is a \texttt{Module}
\newline \textbf{M} contains \textbf{A}
\newline \textbf{N} is a \texttt{Module}
\newline \textbf{N} contains \textbf{B}
\newline \textbf{A} is supported by \textbf{B}
\newline \textbf{M} and \textbf{N} are distinct
  & \textbf{B} is \textit{away} (in another \texttt{Module})
\\ \\

Module \textbf{M} contains \textbf{A}
\newline \texttt{Module} \textbf{N} contains \textbf{B}
\newline \textbf{A} is in context of \textbf{B}
\newline \textbf{M} and \textbf{N} are distinct \texttt{Module}s
  & \textbf{B} is \textit{away} (in another \texttt{Module})
\\ \\

Module \textbf{M} contains both \textbf{A} and \textbf{B}
\newline \textbf{A} and \textbf{B} share the same schema identifier \textbf{ID}
\newline \textbf{A} and \textbf{B} are distinct
  & \textbf{A} is \underline{not} \textit{valid}
\newline \textbf{B} is \underline{not} \textit{valid}
\\ \\

Relationship \textbf{R} has object \texttt{Module} \textbf{M1}
\newline \textbf{M1} and \textbf{M2} are \texttt{Module}s
\newline \textbf{M2} contains \textbf{A}
\newline \textbf{A} is supported by \textbf{M1}
\newline \textbf{R} has subject \textbf{A}
\newline \textbf{M2} is under \textit{contract}
  & \textbf{R} is \underline{not} \textit{valid}
\\ \\

\textbf{A} is in context of \texttt{Module} \textbf{M1}
\newline \texttt{Relationship} \textbf{R} has object \textbf{M1}
\newline \textbf{M1} and \textbf{M2} are \texttt{Module}s
\newline \textbf{M2} contains \textbf{A}
\newline \textbf{R} has subject \textbf{A}
\newline \textbf{M2} is under contract
  & \textbf{R} is \underline{not} \textit{valid}
\\ \\

Module \textbf{M} contains \textbf{A}
\newline \textbf{A} is to be supported by contract
\newline \textbf{M} is under contract
  & \textbf{A} is \underline{not} \textit{valid}
\\ \\

Relationship \textbf{R} links \texttt{Goal} \textbf{G1} to \texttt{Goal} \textbf{G2}
\newline \textbf{M1} and \textbf{M2} are \texttt{Module}s
\newline \textbf{G1} is a \texttt{Goal} contained in \textbf{M1}
\newline \textbf{G2} is a \texttt{Goal} contained in \textbf{M2}
\newline \textbf{G1} is \textit{away} and supported by \textbf{G2}
  & \textbf{R} is \underline{not} \textit{valid}
\\ \\

\textbf{A} is \textit{away} and must be supported by contract
\newline \textbf{M1} and \textbf{M2} are \texttt{Module}s
\newline \textbf{A} is in \textbf{M1}
\newline \textbf{B} is in \textbf{M2}
\newline \textbf{A} is supported by \textbf{B}
\newline \textbf{M1} and \textbf{M2} are distinct
  & \textbf{M2} becomes a \textit{contract} \texttt{Module}
\\ \\

Module \textbf{M} contains both \textbf{A} and \textbf{B}
\newline \textbf{A} is to be supported by contract and is supported by \textbf{B}
\newline \texttt{Relationship} \textbf{R} links \textbf{A} to \textbf{B}
  & \textbf{R} is \underline{not} \textit{valid}
\\ \\

\texttt{Goal} \textbf{G1} is in context of \textbf{C1}
\newline \texttt{Goal} \textbf{G2} is in context of \textbf{C2}
\newline \textbf{G1} and \textbf{G2} are \texttt{Goal}s
\newline \textbf{G2} is \textit{away} and supported by \textbf{G1}
\newline \textbf{C2} is a Context
  & \textbf{C1} is consistent with \textbf{C2}
\\ \\

Module \textbf{M2} contains \texttt{Goal} \textbf{G}
\newline \textbf{M1} and \textbf{M2} are \texttt{Module}s
\newline Justification \textbf{J} is substituted by \textbf{G}
\newline \textbf{M1} contains \textbf{J}
\newline \textbf{J} is a Justification
\newline some element \textbf{E} is in context of \textbf{J}
\newline \textbf{M1} and \textbf{M2} are distinct
  & \textbf{J} is \underline{not} \textit{valid}
\\ \\

\textbf{M1} and \textbf{M2} are \texttt{Module}s
\newline \textbf{E1} in \textbf{M1} is supported by \textbf{E2} in \textbf{M2}
\newline \textbf{M1} and \textbf{M2} are distinct
  & \textbf{M1} is supported by \textbf{M2}
\\ \\

\textbf{M1} and \textbf{M2} are \texttt{Module}s
\newline \textbf{E1} in \textbf{M1} is in context of \textbf{E2} in \textbf{M2}
\newline \textbf{M1} and \textbf{M2} are distinct
  & \textbf{M1} is in context of \textbf{M2}
\\ \\

\textbf{E1} is supported by \texttt{Module} \textbf{M2}
\newline \textbf{M1} and \textbf{M2} are \texttt{Module}s
\newline \textbf{M1} contains \textbf{E1}
\newline \textbf{M1} and \textbf{M2} are distinct
  & \textbf{M1} is supported by \textbf{M2}
\\ \\

\textbf{M1}, \textbf{M2} and \textbf{M3} are \texttt{Module}s
\newline \textbf{M1} is supported by \textbf{M2}
\newline \textbf{M2} is supported by \textbf{M3}
\newline \textbf{M3} is supported by \textbf{M1}
\newline all three \texttt{Module}s are pairwise distinct
  & \textbf{M1} is \underline{not} \textit{valid}
\newline \textbf{M2} is \underline{not} \textit{valid}
\newline \textbf{M3} is \underline{not} \textit{valid}
\\ \\

\textbf{G} is a \textit{public} \texttt{Goal}
\newline \textbf{G} is in context of \textbf{C}
  & \textbf{C} is \textit{public}
\\ \\

\textbf{C} is a \textit{contract}
\newline \textbf{E} is \textit{public}
\newline \textbf{C} contains \textbf{E}
  & \textbf{E} is \underline{not} \textit{valid}
\\ \\

\textbf{A} is an Artefact Reference marked as an \textit{assurance claim point}
\newline (we introduce a reified confidence \texttt{Relationship} \textbf{R} for \textbf{A})
  & \textbf{R} is a \texttt{Relationship} with Confidence whose subject is \textbf{A}
\\ \\

\textbf{A} has schema identifier \textbf{N}
\newline \textbf{B} has schema identifier \textbf{M}
\newline \textbf{A} and \textbf{B} are distinct
\newline \textbf{N} and \textbf{M} are equal
  & \textbf{A} is \underline{not} \textit{valid}
\newline \textbf{B} is \underline{not} \textit{valid}
\\ \\

\texttt{Goal} \textbf{G1} is supported by \texttt{Strategy} \textbf{S}
\newline \textbf{S} is supported by \texttt{Goal} \textbf{G2}
\newline there is a \texttt{Relationship} with Confidence \textbf{R1} from \textbf{G1} to \textbf{S} associated with \textbf{A}
  & we create a \texttt{Relationship} with Confidence \textbf{R2} from \textbf{S} to \textbf{G2}, associated with the same \textbf{A}, and mark \textbf{R2} as an \textit{assurance claim point}
\\ \\

\texttt{Goal} \textbf{G2} is supported by \texttt{Strategy} \textbf{S}
\newline \textbf{S} is supported by \texttt{Goal} \textbf{G1}
\newline there is a \texttt{Relationship} with Confidence \textbf{R1} from \textbf{S} to \textbf{G1} associated with \textbf{A}
  & we create a \texttt{Relationship} with Confidence \textbf{R2} from \textbf{G2} to \textbf{S}, associated with the same \textbf{A}, and mark \textbf{R2} as an \textit{assurance claim point}
\\ \\

\texttt{Goal} \textbf{G} is supported by two distinct \texttt{Solution}s \textbf{S1} and \textbf{S2}
\newline there is a \texttt{Relationship} with Confidence \textbf{R1} from \textbf{G} to \textbf{S1} associated with \textbf{A}
  & we create a \texttt{Relationship} with Confidence \textbf{R2} from \textbf{G} to \textbf{S2}, associated with the same \textbf{A}, and mark \textbf{R2} as an \textit{assurance claim point}
\\ \\

\texttt{Goal} \textbf{G} is supported by two distinct \texttt{Goal}s \textbf{S1} and \textbf{S2}
\newline there is a \texttt{Relationship} with Confidence \textbf{R1} from \textbf{G} to \textbf{S1} associated with \textbf{A}
  & we create a \texttt{Relationship} with Confidence \textbf{R2} from \textbf{G} to \textbf{S2}, associated with the same \textbf{A}, and mark \textbf{R2} as an \textit{assurance claim point}
\\ \\

\textbf{S} is a \textit{valid} \texttt{Solution}
\newline \textbf{S} challenges \textbf{E}
  & \textbf{E} is \textit{defeated}
\\ \\

\textbf{E} is \textit{defeated}
  & \textbf{E} is \underline{not} \textit{valid}
\\ \\

\texttt{Goal} \textbf{G} challenges \textbf{E}
\newline \textbf{G} is \textit{true}
  & \textbf{E} is \textit{in doubt}
\\ \\

\texttt{Goal} \textbf{G} challenges \textbf{E}
\newline \textbf{G} is \underline{not} \textit{undeveloped}
\newline \textbf{E} is \textit{in doubt}
  & \textbf{E} is \textit{defeated}
\\ \\

\textbf{A} challenges \textbf{B}
\newline (we reify this challenge as a \texttt{Relationship} \textbf{R})
  & \textbf{R} is a reified \texttt{Relationship} with subject \textbf{A} and object \textbf{B}
\\ \\

Defeater \textbf{D} challenges \texttt{Relationship} \textbf{R} between \texttt{Goal}s \textbf{G3} (subject) and \textbf{G1} (object)
\newline \textbf{G1} and \textbf{G3} are \texttt{Goal}s
\newline we introduce a \texttt{Strategy} \textbf{S} for \textbf{G3} and another \texttt{Goal} \textbf{G2} distinct from \textbf{G1}
\newline \textbf{G3} is supported by \textbf{G1}
  & \texttt{Strategy} \textbf{S} is supported by \textbf{G1} and \textbf{G2}, \textbf{G3} is supported by \textbf{S}, \textbf{D} now challenges \textbf{S}, and \textbf{D}, \textbf{S} and \textbf{R} are all marked \underline{not} \textit{valid}
\\ \\

\textbf{A} challenges \textbf{B}
  & \textbf{A} is a Defeater
\\ \\

\bottomrule
\caption{Full set of SWRL rules in OntoGSN. Multiple lines within a cell are conjoined with AND.}
\label{tab:swrl-rules-full}
\end{longtable}

%% file: tables/competency_questions.tex
\begin{table}[htbp]
\centering
\renewcommand{\arraystretch}{1.15} 
\begin{tabular}{@{}p{1.5cm}p{10cm}@{}}
\toprule
\textbf{ID} & \textbf{Competency Question} \\
\midrule
\multicolumn{2}{@{}l}{\textbf{Assurance engineer}}\\[0.15em]
AE\textendash01 & Which \textbf{Goals} or \textbf{Strategies} are affected by the \textbf{Defeater} “Jailbreak”? \\
AE\textendash02 & Which \textbf{Solution} instances are flagged \texttt{inDoubt=true}? \\
AE\textendash03 & Which bottom-level \textbf{Goals} in the “Robustness” argument lack any supporting \textbf{Solution}? \\
AE\textendash04 & Which \textbf{Solution} items do not reference any \textbf{Artefact}? \\
AE\textendash05 & Which \textbf{Solution} instances were added more than 180 days ago? \\
\addlinespace[0.35em]
\multicolumn{2}{@{}l}{\textbf{Domain expert}}\\[0.15em]
DE\textendash01 & Which \textbf{Goals} mention “Benchmark” currently lack any associated \textbf{Evidence}? \\
DE\textendash02 & How can listed \textbf{Artefacts} be instantiated as \textbf{Solutions} using the template “Test against \{attack prompt\}”? \\
DE\textendash03 & How can bottom-level \textbf{Goals} with “Perturbation Robustness” be attached to a new \textbf{Artefact} file every 30 days? \\
DE\textendash04 & How can the \textbf{Goal} “Attack Resistance” receive a \textbf{Defeater} once an \textbf{Artefact} with “adversarialSample” is created? \\
DE\textendash05 & Which \textbf{Goals} are in context of a \textbf{Context} that refers to the “GPT” \textbf{Model}? \\
\addlinespace[0.35em]
\multicolumn{2}{@{}l}{\textbf{Auditor}}\\[0.15em]
AU\textendash01 & How can elements in the branch rooted at the \textbf{Goal} “Attack Resistance” be made invalid? \\
AU\textendash02 & How can all bottom-level \textbf{Goals}, \textbf{Solutions}, and \textbf{Artefacts} under the \textbf{Goal} “Attack Resistance” be found? \\
AU\textendash03 & Which elements with “Jailbreak” require a \textbf{Defeater}? \\
AU\textendash04 & How can elements connected to the valid \textbf{Artefact} “BenchmarkDataset” be also marked as \texttt{valid=true}? \\
AU\textendash05 & Which \textbf{Solutions} are attached to a \textbf{ConfidenceArgument} containing a \textbf{Goal} with “Perplexity”? \\
\bottomrule
\end{tabular}
\caption{Stakeholder-specific competency questions, to be supported by OntoGSN.}
\end{table}